# Can Large Language Models Augment a Biomedical Ontology with missing Concepts and Relations?

Antonio Zaitoun[1] [0000-0002-1818-5619], Tomer Sagi[2] [0000-0002-8916-0128], Szymon Wilk[3] [0000-0002-7807-454X], and Mor Peleg[1] [0000-0002-5612-768X]

[1]Department of Information Systems, University of Haifa
azaitoun@campus.haifa.ac.il, morpeleg@is.haifa.ac.il
[2]Department of Computer Science, Aalborg University, Aalborg, Denmark, tsagi@cs.aau.dk
[3]Institute of Computing Science, Poznan University of Technology,
szymon.wilk@cs.put.poznan.pl

**Abstract.** Ontologies play a crucial role in organizing and representing knowledge. However, even current ontologies do not encompass all relevant concepts and relationships. Here, we explore the potential of large language models (LLM) to expand an existing ontology in a semi-automated fashion. We demonstrate our approach on the biomedical ontology SNOMED-CT utilizing semantic relation types from the widely used UMLS semantic network. We propose a method that uses conversational interactions with an LLM to analyze clinical practice guidelines (CPGs) and detect the relationships among the new medical concepts that are not present in SNOMED-CT. Our initial experimentation with the conversational prompts yielded promising preliminary results given a manually generated gold standard, directing our future potential improvements.

## 1 Introduction and Related Work

Ontologies [1] are essential for the organization and management of knowledge within a particular domain. They consist of formalized representations of concepts and the relationships among them. They are widely used in applications such as information retrieval and data integration. In healthcare, ontologies, such as SNOMED-CT [2] are used to standardize medical concepts and relationships among them. However, these ontologies may not always represent the latest developments in medical research and practice [3]. For example, in the European project Cancer Patients Better Life Experience (CAPABLE), we rely on recommendations taken from the cancer-related fatigue guideline [4]. This guideline provides a set of recommendations for the pharmacological and non-pharmacological (physical activity and mindfulness) management of fatigue, including concepts not found in SNOMED-CT. Moreover, SNOMED-CT contains concepts but does not provide a complete ontology with the full set of non is-a relations (attributes) present between therapies, clinical conditions, and other semantic types.

To aid the ontology and knowledge graph completion, many researchers have been using neural network based approaches [5, 6, 19, 20, 21]. These techniques involve identifying concepts from text data (eg, domain relevant documents, drawing relationships between them, and matching the extracted concepts to existing concepts in standard terminologies.

A large language model (LLM) is a neural-network-based AI system that is trained on a substantial corpus of text data to perform natural language processing tasks such as text summarization and question answering [7]. General-purpose LLMs such as BERT [8] and GPT [9] have been extensively studied for their capabilities on language-related tasks. ChatGPT [10], an openly available conversational interface to a GPT LLM, has been used for natural language generation tasks in Chabot systems, allowing them to generate human-like responses. In the medical domain, LLMs have been used for the harmonization of local source concepts to standard clinical terminologies, such as SNOMED-CT [11] and LOINC [12]. However, these works are limited to only some relations (*is-a*). Another motivation for using LLM to extend ontologies is their ability to pack large amounts of knowledge into their large parameter set [13].

In this preliminary work, we examine whether general-purpose LLMs that have not been specifically trained over this task can be utilized for the extension of an ontology and selection of the appropriate relation type. We prompted the LLM with manually extracted concepts from a medical guideline and a set of possible semantic relations to choose from. The results are evaluated by three domain experts.

## 2    Methods

**Gold standard**. Two experts from the CAPABLE project (MP and SW) manually examined 4 subsections from the fatigue guideline [4] discussing non-pharmacological treatments (mindfulness-based stress reduction, yoga, acupuncture, and psychosocial interventions). Working individually, they identified important explicit and implicit relations between domain concepts, and recorded them as triples (`Concept1 relation Concept2. eg, mindfulness-based stress reduction <treats> stress`). These triples could be used to extend SNOMED-CT by adding concepts that are not in SNOMED-CT as well as non is-a relations. We found that SNOMED-CT relations do not capture all of the semantic relations that exist in the UMLS Semantic Network [14] and are important for ontology construction. Therefore, we asked the experts to use the following relation types from the UMLS Semantic Network: is-a, part-of, treats, affects, uses, result-of, contains, manages, disrupts, complicates, interacts-with, and prevents. We then established consensus between the experts on triples to be included in the gold standard.

Note that some of the paragraphs from the fatigue guideline included relationships that were *not* important for the task of extending a domain ontology with new concepts and relations. For example, "Information and counseling can help patients to gain a better understanding of CRF". All these relations were discarded. On the other hand, the experts were asked to add implicit relations, not found explicitly in the text but implied from it.

**ChatGPT prompts**. Our method is summarized in Figure 1. First (step 1), we prompted [15] ChatGPT to identify concepts given the context (fatigue paragraphs [4]). Next (step 2), we prompted ChatGPT to build triples given the context, concepts (identified in the first step), and possible UMLS relations. The prompts are provided in [15]. We included inverse relations due to the nature of LLM being based on the probability of next-word sequence prediction, which may favor certain conjunction forms over others. Hence it may choose to output the inverse rather than the regular relation. For instance, the model might have been trained on text corpus containing the phrase "fatigue is caused by cancer" and therefore may favor the relation *caused-by* over *causes*.

We prompted ChatGPT 10 times (in step 1 and in step 2). Because ChatGPT returns different results when prompted with the same prompt several times, we accepted as results, only the concepts and the triples that appeared in at least 6 of the 10 runs.

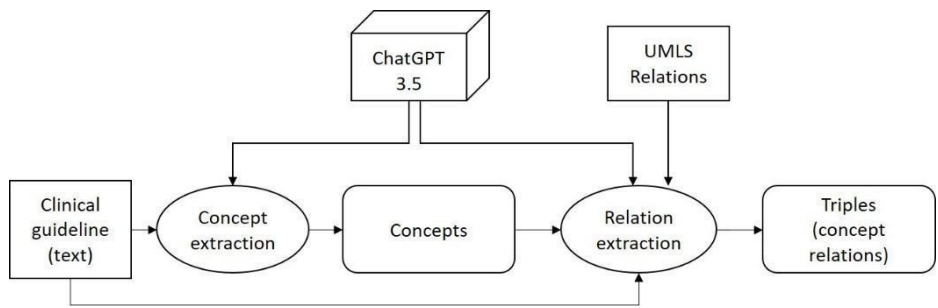

**Fig. 1.** The generation of ontology fragment (triples)

**Post-processing**. In a post-processing step, we reversed the triples that used inverse relation types. The final output, comprising triples, was converted into OWL and visualized using WebVowl [16].

**Evaluation**. We partitioned the triples found by ChatGPT into **4 levels of difficulty** for the LLM and for a human expert:

1) Both concepts are in SNOMED-CT and relation is is-a;
2) At most one concept is in SNOMED-CT and the relation is is-a;
3) Both concepts are in SNOMED-CT and the relation is different from is-a;
4) At most one concept is in SNOMED-CT and relation is not is-a.

Triples found by the LLM and in the gold standard were labeled as *true positives (TP)*, triples not found by the model were treated as *false negatives (FN)*. Triples found by the model but not in the gold standard were reviewed by the two experts who had created the gold standard. They either accepted these triples, in which case they were labeled as *true positives (TP)* and used to extend the gold standard, or declined them, in which case they were labeled as *FP*. These triples play a crucial role in completing the ontology; although they are implicit, they contribute to the creation of a more interconnected ontology. While the humans involved were experts, they could only identify explicit triples, rather than the implicit ones, such as:

- [Stress reduction]-(subClassOf)->[Clinical Intervention]
- [Fatigue]-(subClassOf Physical)->[Functioning]
- [Walking meditation]-(subClassOf)->[Meditation Exercise]

Precision was calculated as TP/(TP+FP) and recall as TP/(TP+FN). We computed precision and recall separately for each of the four levels of difficulty.

## 3   Results

Initially, the gold standard included 49 concepts and 52 relations (triples). ChatGPT found 14 of the relations that were in the gold standard and missed 38 of them. Interestingly, ChatGPT also found 3 concepts and 39 relations that the experts determined to be valid and important for ontology extension, but were not in the original gold standard, hence they were added to the gold standard as TP. Thirty relations found by ChatGPT were judged by the experts to be incorrect. Thus the overall precision is 0.63 and recall 0.58.

29 of the 52 concepts (58%) were included in SNOMED-CT. Moreover, 29 of the 91 extended gold standard relations (32%) were is-a relations but only 9 were between pairs of concepts that were both found in SNOMED-CT. Table 1 presents the results when we partition the triples into the 4 levels of difficulty. The final output is visualized in [17].

Table 1. Precision (prec) and recall for concepts and for relations, considering difficulty level

| Concepts | | | | | Relations | | | | | |
|---|---|---|---|---|---|---|---|---|---|---|
| TP | FN | FP | Prec | Recall | | TP | FN | FP | Prec | Recall |
| 45 | 7 | 3 | 0.93 | 0.86 | Level 1 | 1 | 10 | 11 | 0.08 | 0.08 |
| | | | | | Level 2 | 11 | 14 | 1 | 0.91 | 0.44 |
| | | | | | Level 3 | 36 | 5 | 14 | 0.72 | 0.87 |
| | | | | | Level 4 | 5 | 9 | 4 | 0.55 | 0.35 |

## 4  Discussion

We proposed a method to use general-purpose LLM to infer missing concepts within an existing ontology and assert the appropriate semantic relations between concepts.

The results of our case study suggest that general-purpose LLMs have a potential to be useful tools for extending existing ontologies, specifically as part of a semi-automated process where a human validates the LLM suggestions. However, the precision was only 0.63, which means that many of the relations suggested by the LLM were invalid and the recall was also low (0.58), suggesting that relying entirely on humans for identifying concepts and relations in text may be more complete and correct and perhaps even more efficient than the semi-automated methods that involve human assessment. To try to understand which of the relations suggested by ChatGPT were more complete and correct, we partitioned the relations into 4 levels of difficulty, as explained in the Methods. The results presented in Table 1 are intriguing. We thought that the model would have the highest precision and recall for Level 1 relations. However, the worst results were obtained for this level. The model was actually quite useful in providing specific relations for is-a relations between new concepts (Level 2) and had high recall and fair precision for level 3 relations (non is-a among existing concepts).

In this exploratory study, we employed the default temperature (1.00) setting for generating responses with ChatGPT. The formulation of prompts was an iterative process involving trial and error to optimize their effectiveness by trying including or excluding the context, including the inverse relations, and providing an example triple to ensure a consistent structured output.

It is important to note that the limitations of our study included the use of a single medical guideline, a specific language model (LLM), and a particular sub-domain. To obtain more comprehensive insights, a more robust assessment should involve comparing multiple LLMs across various medical guidelines from diverse domains. Additionally, future experiments should be conducted using updated versions of ChatGPT, as they may yield improved results due to advancements in the underlying models. Exploring LLMs trained on biomedical texts, such as BioMedML (available at https://www.mosaicml.com/blog/introducing-pubmed-gpt), would also be intriguing.

Furthermore, we anticipate that future iterations of ChatGPT would exhibit deterministic behavior, ensuring consistent output for the same input. This would enhance the reliability and usability of the system. To enhance the accuracy of generated relations, further research should be conducted, focusing on incorporating the concept extraction process within the suggestion pipeline, potentially utilizing named-entity-recognition models [18]. These advancements would contribute to the ongoing efforts in refining and expanding the capabilities of ChatGPT.

The results of this case study include concepts and UMLS relations between them that are part of the mental health domain, but are not in SNOMED-CT. We recommend adding them to SNOMED-CT to allow the use of standard terms and relations in knowledge bases of behavior health apps.

## Acknowledgement

The work described has been funded by the European Union's Horizon 2020 research and innovation programme (grant #875052 - CAPABLE).

# 5 Appendix

Table 2. Gold standard table curated by two domain experts.

| Concept A | Relation | Concept B | Classification | In GPT |
|---|---|---|---|---|
| Acupuncture | treats | fatigue | 3 | Yes |
| Cancer-related fatigue | is-a | fatigue | 1 | No |
| cognitive appraisal of CRF | resut-of | Psychosocial intervention | 4 | No |
| cognitive-behavioural interventions | part-of | Mindfulness-based clinical interventions | 4 | No |
| energy conservation | is-a | Psychosocial intervention | 1 | No |
| meditation | part-of | yoga | 1 | No |
| meditation exercise | part-of | mindfulness-based clinical intervention | 4 | No |
| Mindfulness-based stress reduction | affects | well-being | 3 | No |
| Mindfulness-based stress reduction | treats | mental health | 4 | No |
| Mindfulness-based stress reduction | treats | stress | 3 | Yes |
| Mindfulness-based stress reduction | treats | mood | 3 | Yes |
| Mindfulness-based stress reduction | treats | anxiety | 3 | Yes |
| Mindfulness-based stress reduction | treats | depressive mood | 3 | Yes |
| Mindfulness-based stress reduction | treats | sleep | 3 | Yes |
| Mindfulness-based stress reduction | affects | sleep | 3 | Yes |
| Mindfulness-based stress reduction | treats | fatigue | 3 | Yes |
| Mindfulness-based stress reduction | affects | psychological functioning | 3 | Yes |
| Mindfulness-based stress reduction | affects | psychosocial adjustment | 3 | Yes |
| Mindfulness-based stress reduction | affects | stress reduction | 4 | No |
| Mindfulness-based stress reduction | affects | enhancements in coping and well-being symptoms | 4 | No |
| Mindfulness-based stress reduction | affects | quality of life | 3 | No |
| Mindfulness-based stress reduction | treats | fear of recurrence | 4 | No |
| movement exercise | part-of | mindfulness-based clinical intervention | 4 | No |
| physical poses with a focus on breathing | part-of | yoga | 4 | Yes |
| psychoeducation | is-a | Psychosocial intervention | 1 | No |
| psychosocial counselling | is-a | Psychosocial intervention | 1 | No |
| psychosocial interventon | treats | Cancer-related fatigue | 3 | No |
| psychosocial interventon | uses | information communication | 4 | No |
| psychotherapy | is-a | Psychosocial intervention | 1 | No |
| relaxation techniques | is-a | Psychosocial intervention | 1 | No |
| stress management | is-a | Psychosocial intervention | 1 | No |
| yoga | is-a | Mindfulness-based stress reduction | 1 | No |
| yoga | affects | overall QoL | 3 | Yes |
| yoga | treats | fatigue | 3 | No |
| yoga | treats | stress | 3 | No |
| yoga | affects | distress during treatment | 3 | Yes |
| yoga exercise | is-a | Mindfulness-based stress reduction therapy | 1 | No |
| **TOTAL** | | 52 | | |

Table 3. Relationships extracted by ChatGPT

| Subject | relation | Object | In Gold Standard | Type |
|---|---|---|---|---|
| Cancer Patients | subClassOf | Patients | No | 2 |
| Meditation | subClassOf | Yoga | No | 1 |
| Breathing | subClassOf | Yoga | No | 1 |
| Physical Poses | subClassOf | Yoga | No | 2 |
| Yoga | subClassOf | Mind-Body Intervention | Yes | 2 |
| Fear of recurrence | subClassOf | Mental Health | No | 1 |
| Quality of life | subClassOf | Mental Health | No | 1 |
| Enhancements in coping and well-being symptoms | subClassOf | Mental Health | No | 1 |
| Stress reduction | subClassOf | Clinical Intervention | No | 2 |
| Psychosocial adjustment | subClassOf | Mental Health | No | 1 |
| Psychological functioning | subClassOf | Mental Health | No | 1 |
| Fatigue | subClassOf | Physical Functioning | No | 2 |
| Sleep | subClassOf | Physical Functioning | No | 2 |
| Depressive mood | subClassOf | Mental Health | No | 1 |
| Anxiety | subClassOf | Mental Health | No | 1 |
| Mood | subClassOf | Mental Health | No | 1 |
| Stress | subClassOf | Mental Health | No | 1 |
| Insight meditation | subClassOf | Meditation Exercise | No | 2 |
| Walking meditation | subClassOf | Meditation Exercise | No | 2 |
| Yoga exercises | subClassOf | Meditation Exercise | No | 1 |
| Body scan | subClassOf | Meditation Exercise | No | 2 |
| Sitting meditation | subClassOf | Meditation Exercise | No | 2 |
| Mindfulness-based cognitive therapy | subClassOf | Clinical Intervention | No | 2 |
| Mindfulness-based stress reduction | subClassOf | Clinical Intervention | No | 2 |
| Psychosocial interventions | affects | physical activity or training | No | 3 |
| Psychosocial interventions | contains | stress management | No | 3 |
| Psychosocial interventions | contains | energy conservation | No | 3 |
| Psychosocial interventions | contains | relaxation techniques | No | 3 |
| Psychosocial interventions | affects | self-help or self-care strategies | No | 4 |
| Psychosocial interventions | affects | coping strategies and behaviour | No | 3 |
| Psychosocial interventions | affects | cognitive appraisal of CRF | No | 4 |
| Psychosocial interventions | contains | mind-body interventions | No | 4 |
| Psychosocial interventions | contains | psychoeducation | No | 3 |
| Psychosocial interventions | contains | psychotherapy | No | 3 |
| Psychosocial interventions | contains | psychosocial counselling | No | 3 |
| Psychosocial interventions | treats | CRF | No | 3 |
| Acupuncture | disrupts | Adverse Events | No | 3 |
| Acupuncture | treats | CRF | Yes | 3 |
| Fatigue | affects | Cancer | No | 3 |
| Yoga | treats | Cancer-Related Fatigue | No | 3 |
| Yoga | affects | Distress | Yes | 3 |
| Yoga | affects | Stress | No | 3 |
| Yoga | affects | Quality of Life | Yes | 3 |
| Yoga | affects | Fatigue | No | 3 |
| Yoga | affects | Cancer Patients | No | 4 |
| Yoga | part_of | Meditation | No | 3 |
| Yoga | part_of | Breathing | No | 3 |
| Physical Poses | part_of | Yoga | Yes | 4 |
| movement exercises | part_of | Mindfulness-based cognitive therapy | No | 4 |
| cognitive-behavioural interventions | part_of | Mindfulness-based cognitive therapy | No | 3 |
| meditation exercises | part_of | Mindfulness-based cognitive therapy | No | 4 |
| fear of recurrence | result_of | Mindfulness-based stress reduction | No | 3 |
| quality of life | result_of | Mindfulness-based stress reduction | No | 3 |
| enhancements in coping and well-being symptoms | result_of | Mindfulness-based stress reduction | No | 3 |
| stress reduction | result_of | Mindfulness-based stress reduction | No | 3 |
| Mindfulness-based stress reduction | affects | psychosocial adjustment | Yes | 3 |
| Mindfulness-based stress reduction | affects | psychological functioning | Yes | 3 |
| Mindfulness-based stress reduction | affects | sleep | Yes | 3 |
| Mindfulness-based stress reduction | affects | depressive mood | No | 3 |
| Mindfulness-based stress reduction | affects | anxiety | No | 3 |
| Mindfulness-based stress reduction | affects | mood | No | 3 |
| Mindfulness-based stress reduction | affects | stress | No | 3 |
| improvements | result_of | fatigue | No | 4 |
| Mindfulness-based stress reduction | affects | fatigue | No | 3 |
| Mindfulness-based stress reduction | manages | fear of recurrence | No | 3 |
| Mindfulness-based stress reduction | manages | quality of life | No | 3 |
| Mindfulness-based stress reduction | manages | enhancements in coping and well-being symptoms | No | 3 |
| Mindfulness-based stress reduction | manages | stress reduction | No | 3 |
| Mindfulness-based stress reduction | treats | psychosocial adjustment | No | 3 |
| Mindfulness-based stress reduction | treats | psychological functioning | No | 3 |
| Mindfulness-based stress reduction | treats | fatigue | Yes | 3 |
| Mindfulness-based stress reduction | treats | sleep | Yes | 3 |
| Mindfulness-based stress reduction | treats | depressive mood | Yes | 3 |
| Mindfulness-based stress reduction | treats | anxiety | Yes | 3 |
| Mindfulness-based stress reduction | treats | mood | Yes | 3 |
| Mindfulness-based stress reduction | treats | stress | Yes | 3 |
| Mindfulness-based stress reduction | contains | insight meditation | No | 3 |
| Mindfulness-based stress reduction | contains | walking meditation | No | 3 |
| Mindfulness-based stress reduction | contains | yoga exercises | No | 3 |
| Mindfulness-based stress reduction | contains | body scan | No | 3 |
| Mindfulness-based stress reduction | contains | sitting meditation | No | 3 |
| movement exercises | part_of | Mindfulness-based stress reduction | No | 4 |
| cognitive-behavioural interventions | part_of | Mindfulness-based stress reduction | No | 3 |